\documentclass[runningheads]{llncs}
\usepackage{graphicx}
\usepackage{hyperref}

\begin{document}
\title{Radial Prediction Domain Adaption Classifier for the MIDOG 2022 challenge}
\titlerunning{RP-DAC for the MIDOG 2022 challenge}
%
%
\author{Jonas Annuscheit\inst{1}\orcidID{0000-0002-0281-8733} \and
Christian Krumnow\inst{1}\orcidID{0000-0001-9556-8501}}

%
\authorrunning{Jonas Annuscheit et al.}
%
\institute{${}^1$University of Applied Sciences (HTW) Berlin, Center for biomedical image and information processing, Ostendstraße 25, 12459 Berlin, Germany}

%
\maketitle              

\begin{abstract}
This paper describes our contribution to the MIDOG 2022 challenge for detecting mitotic cells. One of the major problems to be addressed in the MIDOG 2022 challenge is the robustness under the natural variance
that appears for real-life data in the histopathology field. 
To address the problem, we use an adapted YOLOv5s model for object detection in conjunction with a new Domain Adaption Classifier (DAC) variant, the Radial-Prediction-DAC, to achieve robustness under domain shifts. In addition, we increase the variability of the available training data using stain augmentation in HED color space. Using the suggested method, we obtain a test set F1-score of 0.6658.


\end{abstract}
\section{Introduction}
Deep Learning can help to reduce the time-consuming parts like counting mitoses on H\&E histology slides and can help to decrease the variability of the annotations \cite{AUBREVILLE2023102699}. In the used dataset of the MIDOG 2022 challenge \cite{midog2022}, unseen tissue types are available in the test dataset, leading to a significant domain shift from the available training set to the test set.

In this work, we adapt a domain adaption classifier (DAC) by replacing the softmax layer with a modified version of the Radial Prediction (RP) Layer introduced in \cite{RPL}.
With our variant we are able to adaptively learn prototypes of each scanner, tissue type, and case id. These prototypes represent the essential collective features of each individual scanner, tissue or case within an abstract embedding space and are then used to remove this information from the detection model. 

\section{Material and Method}

\subsection{Dataset}
The training dataset of the challenge \cite{midog2022} consists of 9501 point annotations of mitotic figures from 403 $2mm^{2}$ regions of H\&E stained tissue from four scanners and five different tissue types, and one tissue type without annotations.

The dataset was split balanced for each tissue type and scanner combination into 80\% for the training set,  10\% for the validation set, and 10\% for an internal test set. For normalization, each $2mm^{2}$ region was resized to the highest resolution available in the dataset, and 30 overlapping patches of 1280x1280 pixels were created. To reduce the number of patches
for each region, only one-third of the patches, with the most annotations of mitotic figures, were used for the training of the network.
The unlabeled data was included in the training process to train the RP-DAC and to remove the domain information from the network.

\subsection{Data Augmentation}
\textbf{Train Time Augmentation:}
To enlarge the source domain, we used the technique introduced by Tellez et al. \cite{tellez2018whole} that scales and shifts the color channels in the HED color space to imitate color variations from different scanners. We use deconvolution \cite{ruifrok2001quantification} to convert an image from RGB to HED, multiply the hematoxylin, eosin, and dab channel, where the latter collects residual parts of the transformation, with three factors ($\alpha_H$, $\alpha_E$, $\alpha_D$), and transform back to RGB.
For each scanner/tissue combination, the values for each of the $\alpha$'s are drawn from a scaled and translated beta distribution that was manually selected. 
Distributions were chosen by first obtaining the interval spanned by
the mean of 100 sampled images of each scanner/tissue combination in HED space in each channel.
We then design for each channel a beta distribution that roughly resembles the distribution of the means of all images in the interval. The means of each scanner/tissue combination typically cover only a part of the interval in each channel. 
The beta distribution was then scaled for each scanner/tissue combination appropriately, such that the mean of the transformed images best resemble the distribution of the means of all images. 
By this, we enlarge the typical colorspace of a given scanner/tissue combination such that it can also resemble the color palette of a different scanner/tissue combination.
Furthermore, we allow for randomly applied horizontal and vertical flips, reflection, up to full rotation, and translation up to $200$ pixels. Furthermore, we randomly apply a weak blur or sharpening on the input image.

\textbf{Test Time Augmentation:} We mirrored the test images to use our trained YOLO model on four image variants. The corresponding four predictions are then combined as described in Sec. \ref{sec:Model and Training} in the evaluation part.

\subsection{Model and Training}
\label{sec:Model and Training}
\textbf{Detection Model}
Our base model for the detection task was the YOLOv5s \cite{glenn_jocher_2020_4154370} model that was pretrained on the COCO dataset. The YOLOv5s model was used because it is a very fast model allowing for object detection with bounding boxes and has a strong community that keeps it updated with modern developments.

The YOLOv5s architecture contains multiple places where the output of two consecutive convolution layers is combined by concatenating the channels. At these places, we introduce individually learnable parameters $p$ that allow reducing the influence of the earlier of the two convolutional layers by scaling its output with $\mathrm{sigmoid}(p)$ for the concatenation. A detailed list of the used layers and the implementation of the model can be found at
\cite{GITLink}.

\begin{figure*}[t]
\centering
\includegraphics[width=\textwidth]{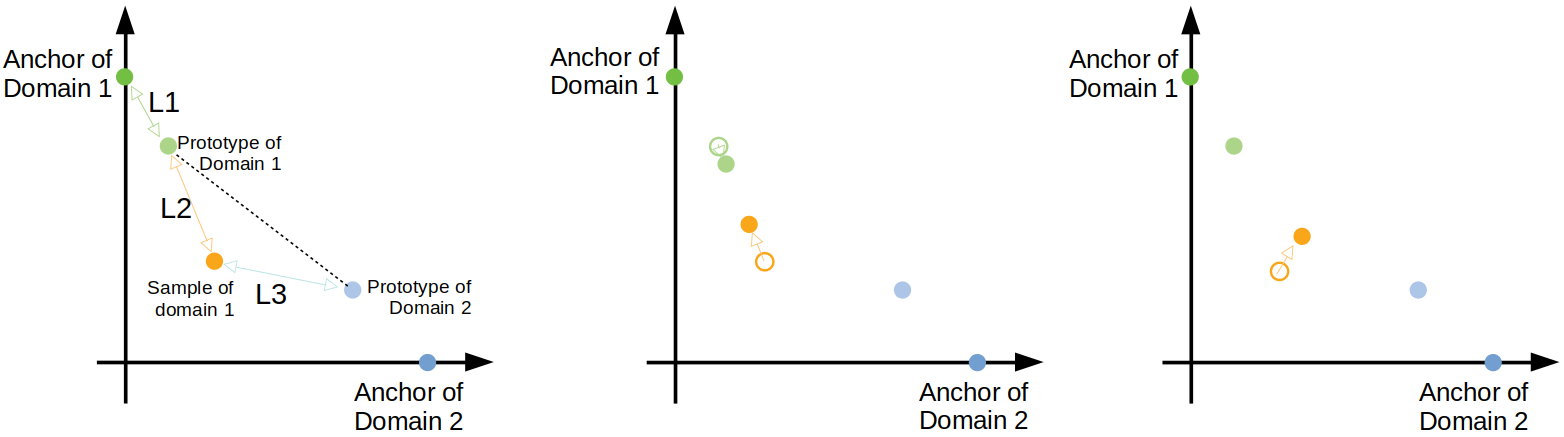}
\caption{
Schematic visualization of the training in the $n$-dimensional space of the RP-DAC for $n=2$. Left panel: initial configuration with two prototypes (green and blue) and one prediction (orange) of an input from domain 1. $L_1$, $L_2$ and $L_3$ denote the distance of the prototype and its anchor or the prediction as indicated. Center panel: during the RP-DAC update (step 1 of the training) distance of prototype and prediction is reduced by minimizing $0.1 L_1^2 + 0.9 L_2^2 $). Right panel: during a YOLO-update step (step 2 of the training ) the weights of the YOLO model are adapted such that the prediction tends towards the center of the prototypes (minimizing the sum $L_2^2+L_3^2$)
}
\label{fig:computerNo}
\end{figure*}

\textbf{RP-DAC:}
In \cite{RPL} the authors developed the Radial Prediction Layer (RPL). For a classification problem with $n$ classes, the RPL is used as a replacement for the softmax layers. It uses $n$ fixed points in an $n$-dimensional space as prototypes (one for each class) and measures the distance of each (embedded) input to each prototype - allowing them to minimize the distance to the prototype of the corresponding class during training.
We modified the idea of the RPL such that it can be used as a DAC with moving prototypes $P_i$ that are linked to fixed anchor $A_i$. Considering $n$ domain classes, we adapt the $n$-dimensional RPL by using the $n$ unit-vectors $\hat{e}_i$ as the fixed anchors $A_i$ and initialize the moving prototypes $P_i$ on the corresponding $n$ with 0.1 scaled unit-vectors, i.e., $P_i = 0.1\,\hat{e}_i$. Each prototype $P_i$ is then linked to the anchor $A_i$ in the training process.

The DAC model itself gets the same input as the detection head of the YOLOv5s model, where we upscale all three feature maps to the largest size occurring and concatenate them along the channels. The DAC then consists of a 1x1 convolution layer that reduces the number of channels to 64, two successive 3x3 convolution layers with stride 2, and a global average pooling. The resulting feature vector is copied and passed to three linear layers that map it into a $n=4$, $n=6$ and $n=403$ dimensional space, respectively ($n=4$ when each scanners, $n=6$ when each tissue types and $n=403$ when each case id constitutes a different domain), followed each by our adapted RP-layer with moving prototype with the corresponding dimension. 
The DAC is trained (in conjunction with the YOLO model as explained below) to classify the different classes in the domain (e.g. identify the scanner of a given image). During that training, the prototypes can move freely in the $n$-dimensional space while the deviation from the corresponding anchor point enters into the loss.

\textbf{Training:}
The training of the whole model is divided into two alternating steps using a batch of $N=64$ accumulated patches (8 mini-batches of size 8) 
and an AdamW optimizer \cite{https://doi.org/10.48550/arxiv.1711.05101} for 800 epochs with learning rates starting at 0.002 and reducing the learning rate by using an OneCycleLR. 
\textbf{1.} A training step for the RP-DACs for a given batch $B$ of inputs $x$ with labels $l_1$, $l_2$ and $l_3$ (where $l_1$ denotes the scanner, $l_2$ the tissue type and $l_3$ the case id of $x$) uses the total loss
$$\mathcal{L}_\mathrm{RP-DAC} = \sum\limits_{i=1,2,3}\frac{1}{N\,n_i}\sum\limits_{(x,l_i)\in B}\left[ 0.1 \cdot \|p_{l}^{(i)}-a_{l}^{(i)}\|_2^2 + 0.9\cdot \|p_{l}^{(i)}- z^{(i)} \|_2^2 \right]$$
with $n_i$ being the dimension, $p_l^{(i)}$ and $a_{l}^{(i)}$ are the prototype and anchor corresponding to the label $l$ and $z^{(i)}$ denotes the prediction for the not augmented input $x$ of the $i$-th RP-DAC. 
Here, only the weights of the RP-DAC layers and the position of the prototypes are updated.
\textbf{2.} A training step for the YOLOv5 model uses the described train-time augmentation and bounding boxes of fixed size around each annotation. The total loss of this training step is given by
$$\mathcal{L}_\mathrm{detection} = \mathcal{L}_{\mathrm{YOLOv5}} + \sum\limits_{i=1,2,3}\frac{1}{N\,n_i}\sum\limits_{l=1}^{n_i} \|p_{l}^{(i)}- z^{(i)} \|_2^2 $$
where $\mathcal{L}_{\mathrm{YOLOv5}}$ denotes the standard YOLOv5 loss and the second term is the averaged distance of the prediction $z^{(i)}$ for $x$ of the $i$-th RP-DAC to all prototypes. Here, only the weights of the YOLOv5 model are updated. The RP-DAC contribution leads the YOLOv5 model to produce representations of the inputs such that their RP-DAC predictions lies in the center of all prototypes, i.e. the YOLOv5 representation is maximally agnostic to the domain (see Fig. \ref{fig:computerNo}).

\textbf{Evaluation:}
During the evaluation, we use the described test time augmentation and obtain for each input image four sets of predictions for bounding boxes of the detected mitosis and corresponding detection probabilities. The four predictions are combined into one set of point annotations by computing the average center of close by bounding boxes. A detected signal is then counted as mitosis if the detection probability is above the threshold of 0.403, which resulted in the best validation set F1-score.

\section{Results}
Our method yields an F1-score of 0.77, 0.74 and 0.75 on our training, validation and internal test set. 
On the not publicly available challenge test set, which contains multiple unseen tissue types we achieve an F1-score of 0.67. 
In order to gauge our method further, we performed an additional training in which the data from the Aperio ScanScope CS2 was left out completely (by this we also left out any patches containing canine cutaneous mast cell tumor in the training). On the data of the left out scanner we then achieve an F1-score of 0.74  while obtaining the F1-scores of 0.81, 0.74 and 0.75 on the corresponding reduced training, validation and test set. From the results of our experiments and on the challenge test set we conclude, that our scheme shows a promising robustness under domain shifts, the limits of which need to investigated in future research.

\section{Discussion}
In this work, we presented our contribution to the MIDOG challenge 2022. Most notably, we introduce the RP-DAC that utilizes moving prototypes instead of fixed classes for a DAC providing additional flexibility to account for instance for overlapping domains. We show that the RP-DAC can be trained stably in conjunction with a YOLOv5s detection model leading to promising domain transferability properties. Learning with such adaptable prototypes provides much potential for future work. Combining prototypes of similar classes using either merging strategies or other clustering methods or using additional domain adaption techniques, e.g., regarding the domain shift removal on residual connections \cite{9506562} are promising routes.

\section{Acknowledgements}
J.A. acknowledges the financial support by the Federal Ministry of Education and Research of
Germany project deep.Health (project number 13FH770IX6).

\bibliographystyle{splncs04}

\begin{thebibliography}{1}
\providecommand{\url}[1]{\texttt{#1}}
\providecommand{\urlprefix}{URL }
\providecommand{\doi}[1]{https://doi.org/#1}

\bibitem{GITLink}
Github repository. \url{https://github.com/JonasAnnuscheit/RPDAC\textunderscore
  FOR \textunderscore MIDOG22}

\bibitem{midog2022}
Aubreville, M., Bertram, C., Breininger, K., Jabari, S., Stathonikos, N., Veta,
  M.: {MI}tosis {DO}main {G}eneralization {C}hallenge 2022. In: 25th
  International Conference on Medical Image Computing and Computer Assisted
  Intervention (MICCAI 2022) (2022). \doi{10.5281/zenodo.6362337}

\bibitem{AUBREVILLE2023102699}
Aubreville, M., Stathonikos, N., Bertram, C.A., Klopfleisch, R., {ter Hoeve},
  N., Ciompi, F., Wilm, F., Marzahl, C., Donovan, T.A., Maier, A., et~al.:
  Mitosis domain generalization in histopathology images — the {MIDOG}
  challenge. Medical Image Analysis  \textbf{84},  102699 (2023).
  \doi{10.1016/j.media.2022.102699}

\bibitem{RPL}
Herta, C., Voigt, B.: Radial {P}rediction {L}ayer. arXiv preprint
  arxiv:1905.11150  (2019). \doi{10.48550/ARXIV.1905.11150}

\bibitem{glenn_jocher_2020_4154370}
Jocher, G.: {ultralytics/yolov5: v3.1 - Bug Fixes and Performance
  Improvements}. \url{https://github.com/ultralytics/yolov5} (Oct 2020).
  \doi{10.5281/zenodo.4154370}

\bibitem{https://doi.org/10.48550/arxiv.1711.05101}
Loshchilov, I., Hutter, F.: Decoupled weight decay regularization. arxiv
  preprint arXiv:1711.05101  (2017). \doi{10.48550/ARXIV.1711.05101}

\bibitem{ruifrok2001quantification}
Ruifrok, A.C., Johnston, D.A.: Quantification of histochemical staining by
  color deconvolution. Analytical and quantitative cytology and histology
  \textbf{23}(4),  291--299 (2001)

\bibitem{tellez2018whole}
Tellez, D., Balkenhol, M., Otte-H{\"o}ller, I., van~de Loo, R., Vogels, R.,
  Bult, P., Wauters, C., Vreuls, W., Mol, S., Karssemeijer, N., et~al.:
  Whole-slide mitosis detection in {H}\&{E} breast histology using {PHH}3 as a
  reference to train distilled stain-invariant convolutional networks. IEEE
  transactions on medical imaging  \textbf{37}(9),  2126--2136 (2018)

\bibitem{9506562}
Zheng, J., Wu, W., Zhao, Y., Fu, H.: Transresnet: Transferable {R}esnet for
  domain adaptation. In: 2021 IEEE International Conference on Image Processing
  (ICIP). pp. 764--768 (2021). \doi{10.1109/ICIP42928.2021.9506562}

\end{thebibliography}

\end{document}